\documentclass[10pt,twocolumn]{article}
\usepackage{simpleConference}
\usepackage{times}
\usepackage{graphicx}
\usepackage{amssymb}
\usepackage[section]{placeins}
\usepackage{float}
\usepackage{amsfonts,amssymb,amsbsy,tabulary,amsmath}
\usepackage{graphicx}
\usepackage{url,multirow,morefloats,floatflt,cancel,tfrupee}
\usepackage{colortbl}
\usepackage{xcolor}
\usepackage{pifont}
\usepackage[nointegrals]{wasysym}
\usepackage[colorlinks,linktocpage]{hyperref}
\hypersetup{
   colorlinks   = true,                               %Colours links instead of ugly boxes
   urlcolor     = blue,                               %Colour for external hyper links
   linkcolor    = blue,                               %Colour of internal links
   citecolor    = red,                                %Colour of citations
   setpagesize  = false,
   linktocpage  = true,
}

\begin{document}

\title{\huge Enhanced Balancing GAN: Minority-class Image Generation}

\author{Gaofeng Huang and Amir H. Jafari \\
\\

The George Washington University, USA \\
\today
\\
\\
ajafari@gwu.edu \\
 ghuang920@gwmail.gwu.edu  \\
}

\maketitle
\thispagestyle{empty}

\begin{abstract}
Generative adversarial networks (GANs) are one of the most powerful generative models, but always require a large and balanced dataset to train. Traditional GANs are not applicable to generate minority-class images in a highly imbalanced dataset. Balancing GAN (BAGAN) is proposed to mitigate this problem, but it is unstable when images in different classes look similar, e.g. flowers and cells. In this work, we propose a supervised autoencoder with an intermediate embedding model to disperse the labeled latent vectors. With the improved autoencoder initialization, we also build an architecture of BAGAN with gradient penalty (BAGAN-GP). Our proposed model overcomes the unstable issue in original BAGAN and converges faster to high quality generations. Our model achieves high performance on the imbalanced scale-down version of MNIST Fashion, CIFAR-10, and one small-scale medical image dataset. \footnote{\url{https://github.com/GH920/improved-bagan-gp}}\mbox{~}
\end{abstract}

\section{Introduction}
Image classification is a classical topic in computer vision. There are many state-of-the-art networks proposed in the ImageNet challenge \unskip~\cite{850219:19990866}. These deep neural networks commonly require a large and balanced dataset for training. However, in medical image classification, the performance of most networks will deteriorate due to the imbalanced dataset. The underlying idea of neural networks is minimizing the loss function via gradient descent. When training on an imbalanced dataset, the gradients will easily fall into the trap of predicting majority. Apart from reducing majority-class samples, to the best of our knowledge, the only effective solution is increasing the samples of minority. In the field of medical images, collecting pathological cases is time-consuming. The best solution is generating new minority-class images with high quality and with diversity.

Generative adversarial networks (GANs) \unskip~\cite{850219:19990867} are currently the most powerful generative models. As one of deep neural networks, GANs also require a large dataset for training. However, the minority-class subset is always insufficient to train a good GAN. In particular, balancing GAN (BAGAN) \unskip~\cite{850219:19990887} provided a new method to train GANs on imbalanced datasets while specifically aiming to generate minority-class images in high quality. The main contributions of BAGAN are: 1. using an autoencoder to initialize the GAN training, which gives the GAN a common knowledge of all classes, 2. combining real/fake loss and classification loss fairly into one output at the discriminator, which ensures a balanced training for each class.

Although BAGAN proposed an autoencoder initialization to stabilize the GAN training, sometimes the performance of BAGAN is still unstable especially on medical image datasets. Medical image datasets are always: 1. highly imbalanced due to the rare pathological cases, 2. hard to distinguish the difference among classes. As shown in \unskip~\cite{850219:19990887}, the imbalanced \textit{Flowers} dateset has many similar classes so that BAGAN performs not well. In our experiments, BAGAN fails to generate good samples on a small-scale medical image dataset. We consider that the encoder fails to separate images by class when translating them into latent vectors. Furthermore, similar to traditional GANs, BAGAN is hard to train and sensitive to its architecture and hyperparameters. Our objective of this work is to generate minority-class images in high quality even with a small-scale imbalanced dataset. Our contributions are:

\begin{itemize}
  \item \relax We improve the loss function of BAGAN with gradient penalty and build the corresponding architecture of generator and discriminator (BAGAN-GP).
  \item \relax We propose a novel architecture of autoencoder with an intermediate embedding model, which helps the autoencoder learn the label information directly.
  \item \relax We discuss the drawbacks of the original BAGAN and exemplify performance improvements over the original BAGAN and demonstrate the potential reasons.
\end{itemize}

\section{Literature review}

Generative adversarial networks (GANs)\unskip~\cite{850219:19990867,850219:19992469} is a minimax problem, which is one of zero-sum non-cooperative games. A typical GAN model contains a generator and a discriminator. The generator wants to maximize its performance, which works to generate images as real as possible to confuse the discriminator. The discriminator works to distinguish a mixture of original and generated images whether real or fake. In this game, the generator attempts to mimic the distribution of the real data.

GAN techniques are fast developed in recent years. There are various types of GANs: with different metrics of comparing two distributions (e.g. KL divergence for the original GAN \unskip~\cite{850219:19990867}, Wasserstein distance for WGAN \unskip~\cite{850219:19990881,850219:19990865}, EBGAN \unskip~\cite{850219:19990880}, BEGAN \unskip~\cite{850219:19990874}, Loss-Sensitive GAN\unskip~\cite{850219:19992560}), with regularization on the loss function (e.g. WGAN-GP\unskip~\cite{850219:19990881}, DRAGAN \unskip~\cite{850219:19990868}), with different well-designed architecture of GANs (e.g. CycleGAN \unskip~\cite{850219:19990879,850219:19990885}, PGGAN \unskip~\cite{850219:19990870}, SAGAN \unskip~\cite{850219:19990875}), with using a single image for generation (e.g. SinGAN \unskip~\cite{850219:19990872}), with conditions (e.g. ACGAN \unskip~\cite{850219:19990886}), for augmentation (e.g. AugGAN \unskip~\cite{850219:19990876}, BAGAN \unskip~\cite{850219:19990887}), for reducing mode collapse problem (e.g. VEEGAN \unskip~\cite{850219:19990873}).

Data augmentation can extract more information from the original datasets to improve the performance of models. Traditional image augmentation is simply applying linear transformations to the original images, e.g. reflections, rotations and shears. If the linear transformations do not affect the recognition of images, it is effective for the models to learn more information on the original dataset. To extract more information, it is also reasonable to apply some non-linear transformations to the original dataset. GANs are exactly good at create similar images by non-linear transformations inside the network. The literature review\unskip~\cite{850219:19990869} compared many data augmentation methods in deep learning, especially the methods based on GANs.

GANs can simulate the distribution of the real dataset and generate new data samples with high quality. Therefore, there are some recent work applying GANs as an augmentation technique. However, the small training set of minority-class images is still a challenge to train a GAN to generate high quality samples. AugGAN\unskip~\cite{850219:19990876} and AugCGAN \unskip~\cite{850219:19990885} proposed an image-to-image translation framework to generate images in target domain. BAGAN\unskip~\cite{850219:19990887} proposed an overall approach to generate minority-class images with high quality to balance the original dataset. \unskip~\cite{850219:19990883} used conditional WGAN-GP (cWGAN-GP) to generate face emotion samples for data augmentation. \unskip~\cite{850219:19990878} discussed the importance of data augmentation in medical image analysis and considered GANs as the most promising technique. For brain tumor images synthesis,\unskip~\cite{850219:19990882} used GANs and\unskip~\cite{850219:20006965} used conditional PGGAN for better tumor detection.

\section{Methods}

\subsection{BAGAN architecture.}

\begin{figure}[htbp]
\centering
\includegraphics[width=3.5in, ,height= 1.8in]{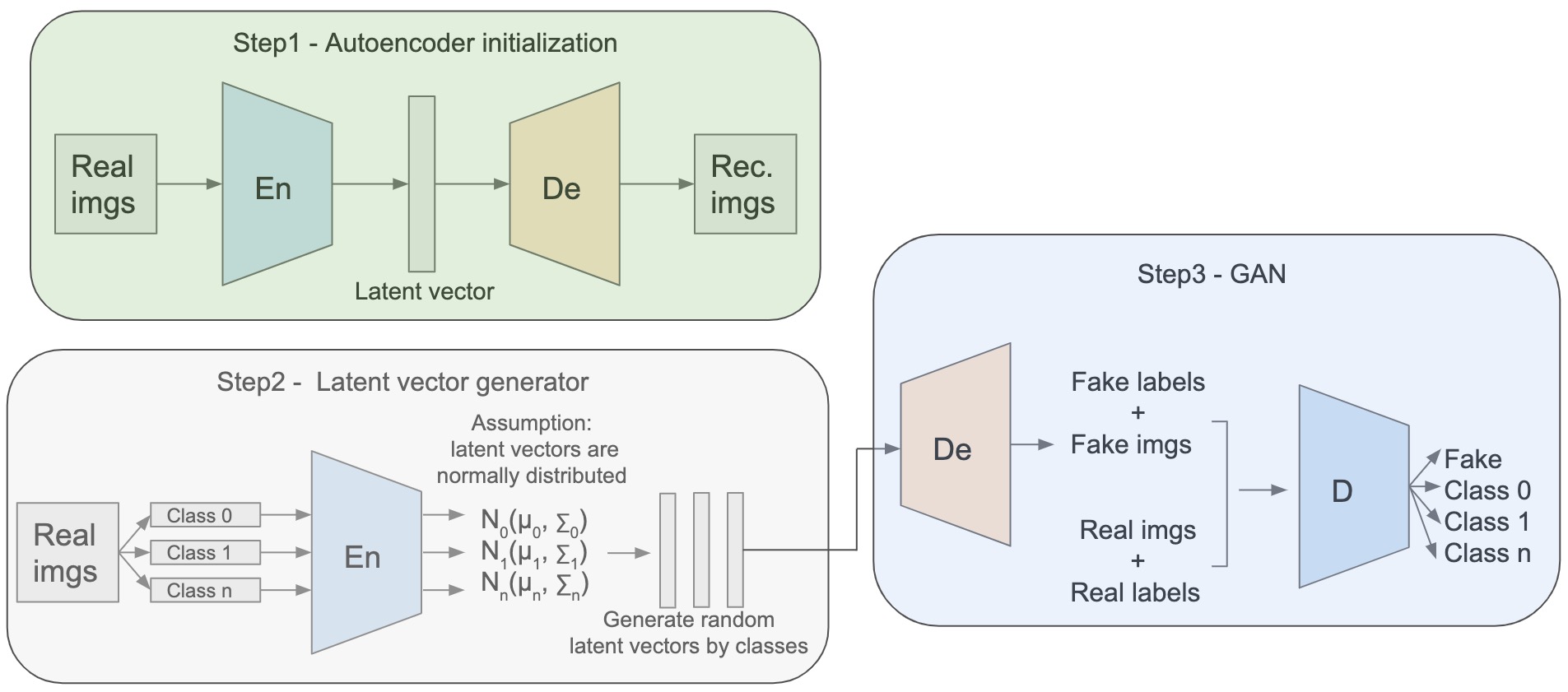}
\caption{\textmd{The architecture of BAGAN. BAGAN proposed three effective steps to improve the quality of generated images when training GANs on imbalanced datasets.}}
\label{f-81eb485cabc0}
\end{figure}

Autoencoder initialization helps generator and discriminator to build a common knowledge of the dataset among all classes. Besides, autoencoder will lead the initialized GAN to a good and stable solution. BAGAN uses a typical autoencoder, the encoder translates a given image into a latent vector and the decoder translates a given latent vector back to a reconstructed image. It applies  \textit{L2} loss minimization between real images and reconstructed images to train the autoencoder networks. In this step, there is no information about classes and the autoencoder learns all images unsupervisedly.

for Labeled latent vectors generation the class information is attached to each latent vector. The real images can be divided into different classes. Using the encoder to translate these images into latent vectors. With an assumption that these latent vectors are normally distributed within their own classes, a probabilistic generator can be derived by calculating means and covariances w.r.t classes.

The generator and the discriminator have prior knowledge from the initialized autoencoder. The generator inherits the same architecture and weights from the trained decoder. The discriminator inherits the same weights of the trained encoder as the first part and adds an auxiliary \textit{softmax} layer to identify different classes. Differently from ACGAN \unskip~\cite{850219:19990886}, the discriminator has only one output but it can classify real/fake and other real classes. Furthermore, in each training batch, the proportion of fake images is the same as any other class. It means the gradients propagated equally for each class and real/fake validity. Although the majority-class images are easier for GAN to learn and to generate real-like images, the balanced training guarantees that the minority-class images will not be ignored.

\subsection{Improvements on BAGAN}

\subsubsection{Improved loss function}

In this work, we will use two advanced loss functions with gradient penalty (from WGAN-GP \unskip~\cite{850219:19990881}  and DRAGAN \unskip~\cite{850219:19990868} ) to compare against the original loss function of BAGAN.

In original GAN model, the loss function is based on KL-JS divergence. Using cross-entropy loss to minimize the difference between two distribution is equivalent to minimizing the KL-JS divergence. However, KL-JS divergence can only give meaningful gradients when two distributions have overlaps. KL-JS divergence cannot measure how far two distributions away when they have no intersections. The loss function $L\left(X_r,X_g\right) $ of original GAN is defined as:

\begin{align*}
\underset{\theta_G}{\min}\underset{\theta_D}{\max}L\left(X_r,X_g\right)&=\mathbb{E}_{x_r\sim X_r}\left[\log\left(D\left(x_r\right)\right)\right]\\
&+\mathbb{E}_{x_g\sim X_g}\left[\log\left(1-D\left(x_g\right)\right)\right]
\tag{1}\label{disp-formula-group-3f359d86ab11448abc26cc2abf311f65}\end{align*}
where \textit{D} denotes the discriminator function, \textit{G} denotes the generator function, $\theta_G $ is the parameters of the generator, $\theta_D $ is the parameters of the discriminator; $x_r $ is sampled from the real distribution $X_r $, $x_g $ is sampled from the generated distribution $X_g $, where $x_g=G\left(z\right) $ and z~is a random noise vector sample from normal distribution$z\sim N\left(0,I_{dim(z)}\right) $. The discriminator is minimizing:

\begin{align*}
L^{\left(D\right)}\left(X_r,X_g\right)=&-\mathbb{E}_{x_r\sim X_r}\left[\log\left(D\left(x_r\right)\right)\right]\\
&-\mathbb{E}_{x_g\sim X_g}\left[\log\left(1-D\left(x_g\right)\right)\right]
\tag{2}\label{disp-formula-group-80a670e90fef473eb3ca7820940ef4ff}\end{align*}
The generator is minimizing:
\let\saveeqnno\theequation
\let\savefrac\frac
\def\dispfrac{\displaystyle\savefrac}
\begin{eqnarray}
\let\frac\dispfrac
\gdef\theequation{3}
\let\theHequation\theequation
\label{disp-formula-group-201e0e71babb400987b6fe5db480bcda}
\begin{array}{@{}l}L^{\left(G\right)}\left(X_g\right)=-\mathbb{E}_{x_g\sim X_g}\left[\log\left(D\left(x_g\right)\right)\right]\end{array}
\end{eqnarray}
\global\let\theequation\saveeqnno
\addtocounter{equation}{-1}\ignorespaces

For the loss function in WGAN, we can replace the KL divergence by the Wasserstein distance to improve the performance and training stability. In practice of constructing an original GAN, the architecture of discriminator is not suggested to be very powerful. A powerful discriminator cannot give meaningful gradients when training its generator. WGAN\unskip~\cite{850219:19990865} proposed the Wasserstein distance to solve this problem. Wasserstein distance is the minimum transport cost of moving mass from one distribution to another distribution, which is also called as Earth-Mover Distance (EMD). EMD is continuous and differentiable so that the gradients are always meaningful, which ensures the stability of the GAN training. Based on the theory of WGAN, the generator will eventually converge to the performance of the discriminator. Hence, WGAN requires a deep architecture of the discriminator so that the discriminator can reach the optimal critic performance. The EMD is defined as:
\let\saveeqnno\theequation
\let\savefrac\frac
\def\dispfrac{\displaystyle\savefrac}
\begin{eqnarray}
\let\frac\dispfrac
\gdef\theequation{4}
\let\theHequation\theequation
\label{disp-formula-group-5ee45758e6f4415aaf9d79a2f4ebd608}
\begin{array}{@{}l}W\left(X_r,X_g\right)=\underset{\gamma\sim\Pi\left(X_r,X_g\right)}{\inf}\mathbb{E}_{\left(x_r,x_g\right)\sim\gamma}\ensuremath{\Vert}{x_r-x_g}\ensuremath{\Vert}\end{array}
\end{eqnarray}
\global\let\theequation\saveeqnno
\addtocounter{equation}{-1}\ignorespaces
where $\Pi\left(X_r,X_g\right) $ denotes all possible joint distributions between the real distribution $X_r $ and the generated distribution $X_g $. Each $\gamma $ represents a transport plan.

However, it is impossible to find the lower bound by traversing all the possible $\gamma $ in this equation. Using the Kantorovich-Rubinstein duality, it is equivalent to find the upper bound in:
\begin{align*}
W\left(X_r,X_g\right)=\underset{\ensuremath{\Vert}D\ensuremath{\Vert}_L\leq1}{\sup}
\left(\mathbb{E}_{x_r\sim{X_r}}\left[D(x_r)\right]
-\mathbb{E}_{x_g\sim X_g}\left[D(x_g)\right]\right)\\
&
\tag{5}\label{disp-formula-group-fc6790da098f4a75a2a59a4ede032a69}\end{align*}
where $\ensuremath{\Vert}D\ensuremath{\Vert}_L\leq1 $ denotes  \textit{D} belongs to the set of 1-Lipschitz functions. Without the constraint, the objective function for the discriminator  is maximizing:
\let\saveeqnno\theequation
\let\savefrac\frac
\def\dispfrac{\displaystyle\savefrac}
\begin{eqnarray}
\let\frac\dispfrac
\gdef\theequation{6}
\let\theHequation\theequation
\label{disp-formula-group-be608968cb5e4c5ea9b28c0ef1ef6195}
\begin{array}{@{}l}W^{\left(D\right)}\left(X_r,X_g\right)=\mathbb{E}_{x_r\sim{X_r}}\left[D\left(x_r\right)\right]-\mathbb{E}_{x_g\sim{X_g}}\left[D\left(x_g\right)\right]\end{array}
\end{eqnarray}
\global\let\theequation\saveeqnno
\addtocounter{equation}{-1}\ignorespaces
The discriminator in WGAN uses an unconstrained real number rather than a classification probability to measure the validity of real/fake images. The loss function of the WGAN does not have a \textit{log-sigmoid} functions comparing to the original GAN.

In Gradient penalty, 1-Lipschitz constraint is equivalent to the norm of gradients $\ensuremath{\Vert}\nabla_{{x}}{D}({x})\ensuremath{\Vert}_2\leq 1 $ everywhere. The gradient penalty term is defined as:
\let\saveeqnno\theequation
\let\savefrac\frac
\def\dispfrac{\displaystyle\savefrac}
\begin{eqnarray}
\let\frac\dispfrac
\gdef\theequation{7}
\let\theHequation\theequation
\label{disp-formula-group-0eb876cf7df6435bb9393c34032b5252}
\begin{array}{@{}l}GP=\mathbb{E}_{{x}\sim{{X}}}\left[(\ensuremath{\Vert}\nabla_{{x}}{D}({x})\ensuremath{\Vert}_2-1)^{2}\right]\end{array}
\end{eqnarray}
\global\let\theequation\saveeqnno
\addtocounter{equation}{-1}\ignorespaces
In WGAN-GP \unskip~\cite{850219:19990881}, they add an extra gradient penalty term to the discriminator loss function. The loss function for the discriminator is minimizing:
\begin{align*}
W^{\left(D\right)}\left(X_r,X_g\right)=
\mathbb{E}_{x_r\sim{X_r}}\left[D\left(x_r\right)\right]-\mathbb{E}_{x_g\sim{X_g}}\left[D\left(x_g\right)\right]\\
+\lambda\mathbb{E}_{\hat{x}\sim{\hat{X}}}\left[(\ensuremath{\Vert}\nabla_{\hat{x}}{D}(\hat{x})\ensuremath{\Vert}_2-1)^{2}\right]
\tag{8}\label{disp-formula-group-ed757d72501d466ab905ece38df8c3c2}\end{align*}
where $\widehat x=\alpha x_r+\left(1-\alpha\right)x_g,\alpha\sim U(0,1) $, which we refer to as ``model interpolation,'' $\lambda $ is a hyperparameter of the penalty extent.

Gradient penalty is only applied in the discriminator loss. The loss function for generator is minimizing:
\let\saveeqnno\theequation
\let\savefrac\frac
\def\dispfrac{\displaystyle\savefrac}
\begin{eqnarray}
\let\frac\dispfrac
\gdef\theequation{9}
\let\theHequation\theequation
\label{disp-formula-group-1da88a5f449842e98366d7e7c9c539c5}
\begin{array}{@{}l}W^{\left(G\right)}\left(X_g\right)=-\mathbb{E}_{x_g\sim{X_g}}\left[D\left(x_g\right)\right]\end{array}
\end{eqnarray}
\global\let\theequation\saveeqnno
\addtocounter{equation}{-1}\ignorespaces
DRAGAN\unskip~\cite{850219:19990868} borrowed the idea of gradient penalty from WGAN-GP\unskip~\cite{850219:19990881} . \unskip~\cite{850219:19990881} indicated the gradient penalty term can be adapted to standard GAN loss function Equation~\ref{disp-formula-group-3f359d86ab11448abc26cc2abf311f65}. \unskip~\cite{850219:19990868} applied the gradient penalty based on the Wasserstein distance to the original \textit{log-sigmoid} loss function and \unskip~\cite{850219:20000466}  demonstrated it is also effective. The loss function for the discriminator is minimizing:

\begin{align*}
L^{\left(D\right)}\left(X_r,X_g\right)=&-\mathbb{E}_{x_r\sim X_r}\left[\log\left(D\left(x_r\right)\right)\right]\\
&-\mathbb{E}_{x_g\sim X_g}\left[\log\left(1-D\left(x_g\right)\right)\right]\\
&+\lambda\mathbb{E}_{\hat{x}\sim{\hat{X}}}\left[(\ensuremath{\Vert}\nabla_{\hat{x}}{D}(\hat{x})\ensuremath{\Vert}_2-1)^{2}\right]
\tag{10}\label{disp-formula-group-2a53c1ae28e74027a6e961e00f5fa0f7}\end{align*}

where $\widehat x=\alpha x_r+\left(1-\alpha\right)x_{noise},\alpha\sim U(0,1),x_{noise}\sim p_{noise} $, which we refer to as ``noise interpolation.'' Although DRAGAN modified the gradient penalty comparing with WGAN-GP, we will not discuss deeply on the difference.

There is no gradient penalty in the generator loss, so the loss function is the same as the original GAN:
\let\saveeqnno\theequation
\let\savefrac\frac
\def\dispfrac{\displaystyle\savefrac}
\begin{eqnarray}
\let\frac\dispfrac
\gdef\theequation{11}
\let\theHequation\theequation
\label{disp-formula-group-ad02331bb7104248973dc689b8ac3b7e}
\begin{array}{@{}l}L^{\left(G\right)}\left(X_g\right)=-\mathbb{E}_{x_g\sim X_g}\left[\log\left(D\left(x_g\right)\right)\right]\end{array}
\end{eqnarray}
\global\let\theequation\saveeqnno
\addtocounter{equation}{-1}\ignorespaces
With comparison of these loss functions in practice, our improved BAGAN uses a DRAGAN-like loss function with the ``model interpolation'' gradient penalty.

For data augmentation, we need to apply conditional GAN to generate minority-class samples. The architecture of DRAGAN and WGAN-GP are almost the same. Referring to ACGAN\unskip~\cite{850219:19990886} and cWGAN-GP \unskip~\cite{850219:19990883}, we built a feasible architecture for conditional DRAGAN (cDRAGAN). Due to the existence of gradient penalty, we cannot add \textit{softmax} layer to the end of the discriminator to identify different classes. The output of the discriminator still needs to be an unconstrained real number. In our work, we keep the output of the generator and the discriminator the same as WGAN-GP whereas we attach the label information into the input of the generator and the discriminator. The label information is expanded by an \textit{embedding} layer and combined with other inputs by a \textit{multiply} layer.
The loss function for the discriminator:

\begin{align*}
L^{\left(D\right)}\left(X_r,X_g,Y_r\right)=-\mathbb{E}_{(x_r, y_r)\sim (X_r,Y_r)}\left[\log\left(D\left(x_r,y_r\right)\right)\right]\\
-\mathbb{E}_{(x_g,y_r)\sim (X_g,Y_r)}\left[\log\left(1-D\left(x_g,y_r\right)\right)\right]\\
+\lambda\mathbb{E}_{(\hat{x},y_r)\sim({\hat{X},Y_r})}\left[(\ensuremath{\Vert}\nabla_{(\hat{x},y_r)}{D}(\hat{x},y_r)\ensuremath{\Vert}_2-1)^{2}\right]
\tag{12}\label{disp-formula-group-ec7ed792a3c14dc8b343d2549f6bfa7f}\end{align*}
Similar to ACGAN and cWGAN-GP, the generated images use the real labels for training in both \textit{G} and \textit{D}.
The loss function for the generator:

\let\saveeqnno\theequation
\let\savefrac\frac
\def\dispfrac{\displaystyle\savefrac}
\begin{eqnarray}
\let\frac\dispfrac
\gdef\theequation{13}
\let\theHequation\theequation
\label{disp-formula-group-c84b057286684ce5a018f2db7df4e779}
\begin{array}{@{}l}L^{\left(G\right)}\left(X_g,Y_r\right)=-\mathbb{E}_{(x_g,y_r)\sim (X_g,Y_r)}\left[\log\left(D\left(x_g,y_r\right)\right)\right]\end{array}
\end{eqnarray}
\global\let\theequation\saveeqnno
\addtocounter{equation}{-1}\ignorespaces

BAGAN has state-of-the-art performance of generating minority-class images on imbalanced datasets. The GAN architecture in BAGAN is just a typical conditional GAN. We improved the GAN part in BAGAN by adopting the architecture and loss function from the cDRAGAN proposed in the previous section. The loss function is modified by the idea of balanced training from BAGAN. The loss function of the discriminator:

\begin{align*}
L^{\left(D\right)}\left(X_r,Z,Y_r,Y_f,Y_{wrong}\right)=\\
-\mathbb{E}_{(x_r, y_r)\sim (X_r,Y_r)}\left[\log\left(D\left(x_r,y_r\right)\right)\right]\\
-\mathbb{E}_{(z,y_f)\sim (Z,Y_f)}\left[\log\left(1-D\left(G(z,y_f),y_f\right)\right)\right]\\
-\mathbb{E}_{(x_r, y_{wrong})\sim (X_r,Y_{wrong})}\left[\log\left(1-D\left(x_r,y_{wrong}\right)\right)\right]\\
+\lambda\mathbb{E}_{(\hat{x},y_r)\sim({\hat{X},Y_r})}\left[(\ensuremath{\Vert}\nabla_{(\hat{x},y_r)}{D}(\hat{x},y_r)\ensuremath{\Vert}_2-1)^{2}\right]
\tag{14}\label{disp-formula-group-d1a9db7736c7485890c7d108e4251a74}\end{align*}
where $z $ is a random noise vector $z\sim N\left(0,I_{\dim\left(z\right)}\right)\equiv Z~$, $y_{f}\sim U\{0,1,2,...\}\equiv Y_{f} $ and$\;y_{wrong}\sim U\{0,1,2,...\;\}\equiv Y_{wrong} $. Previously, the real labels are shared with the real images and the fake images when training the discriminator. In an imbalanced dataset, the real labels randomly sampled from the dataset are still imbalanced. Hence, the GAN will automatically train more on the majority classes. In practice, if we sample from the stratified real labels for training, the GAN will learn slowly. Referring to BAGAN, we randomly sample a fake label from a balanced-label set $Y_{f} $ for each fake image. In order to enhance the learning of class information from the real dataset, we add an extra cross-entropy loss of wrongly classified cases. For the gradient penalty term, we borrow the ``model interpolation'' method from WGAN-GP.

In the setting of balanced training, the loss function of the generator becomes:
\let\saveeqnno\theequation
\let\savefrac\frac
\def\dispfrac{\displaystyle\savefrac}
\begin{eqnarray}
\let\frac\dispfrac
\gdef\theequation{15}
\let\theHequation\theequation
\label{disp-formula-group-b72c8c17c6e648ca98ee6b1c48a7d2ce}
\begin{array}{@{}l}L^{\left(G\right)}\left(Z,Y_{f}\right)=-\mathbb{E}\left[\log\left(D\left(G\left(z,y_{f}\right)\right)\right)\right]\end{array}
\end{eqnarray}
\global\let\theequation\saveeqnno
\addtocounter{equation}{-1}\ignorespaces

\subsubsection{Improved autoencoder}
BAGAN has two key steps comparing with ordinary conditional GAN: autoencoder initialization and labeled latent generation. In our work, we design a new autoencoder architecture with an embedding section. In BAGAN, the labeled latent generation is based on the assumption that the latent vectors are normally distributed. This assumption restricts the performance of BAGAN in practice.

\begin{figure}[htbp]
\centering
\includegraphics[width=2.5in, ,height= 2.2in]{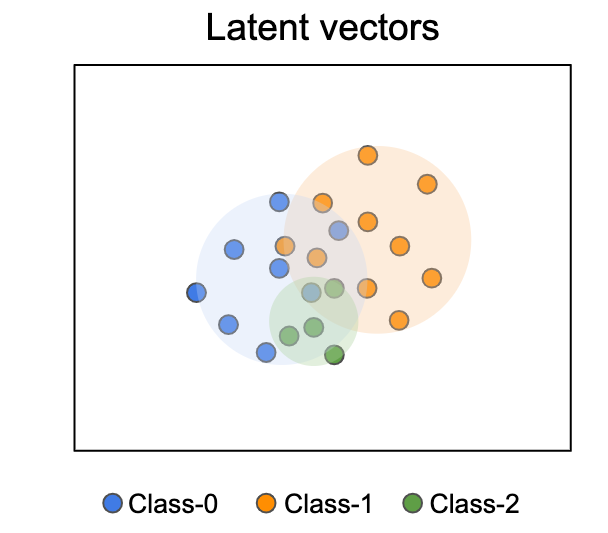}
\caption{\textmd{Distributions of latent vectors in different classes are overlapped}}
\label{figure-5e3299c79bf7452090d158abf1671e6a}
\end{figure}

1. There might be some overlaps between the latent-vector distributions of different classes Figure~\ref{figure-5e3299c79bf7452090d158abf1671e6a} . The result is the generated samples based on the intersected latent vectors look like the mixed-class images. In application, we cannot feed a random latent vector into generator to get images by class. Instead, we must calculate a labeled latent vector by means and covariances of encoded training data.

2. The autoencoder does not learn the label information directly in BAGAN. The latent vectors encoded by the autoencoder cannot disperse their own classes. The labeled latent vectors are defined and restricted by their overlapped distributions, i.e. the label information is unclear. Then, the rough label information attached to the latent vectors will mislead the later GAN training. Furthermore, even if we have a perfectly dispersed latent vectors, the labeled latent vectors are only suitable to the trained decoder. Along with the GAN training, the generator (pretrained decoder) will be updated. However, after the autoencoder initialization, the distributions of labeled latent vectors cannot be updated anymore when we train the later GAN model. In our work, we use an embedding model to generate labeled latent vectors.

\begin{figure}[htbp]
\centering
\includegraphics[width=2.5in, ,height= 1.2in]{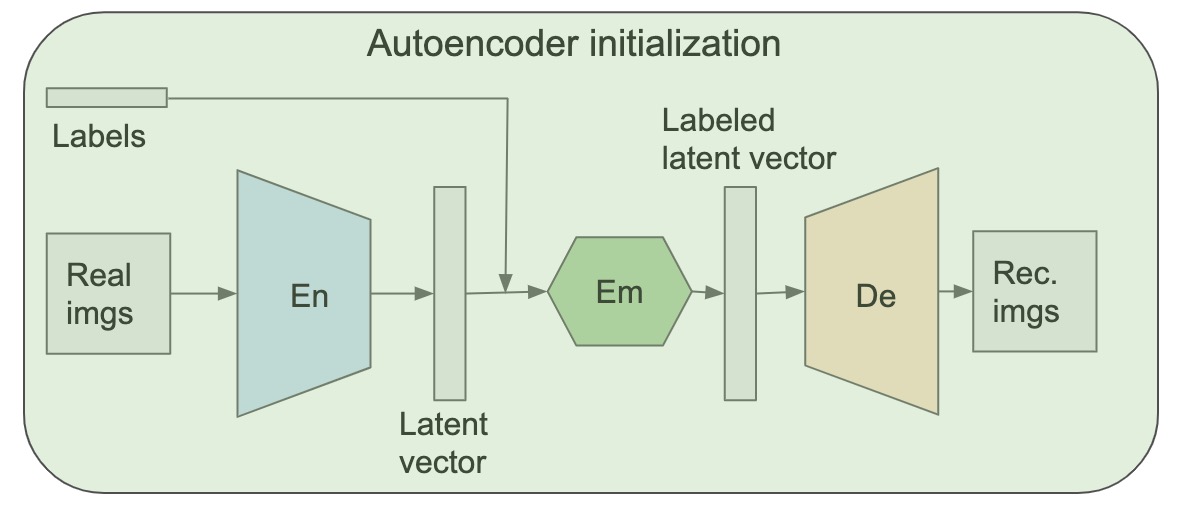}
\caption{\textmd{Autoencoder with an intermediate embedding model.}}
\label{figure-b64e5e7cbe4e4619aec95b9776892e78}
\end{figure}

Our proposed autoencoder is supervised. The label information is embedded to a dense vector with the same size of the latent vector. Then, we apply a \textit{multiply} layer to combine these two vectors as a labeled latent vector.

\begin{figure}[htbp]
\centering
\includegraphics[width=2.5in, ,height= 1.2in]{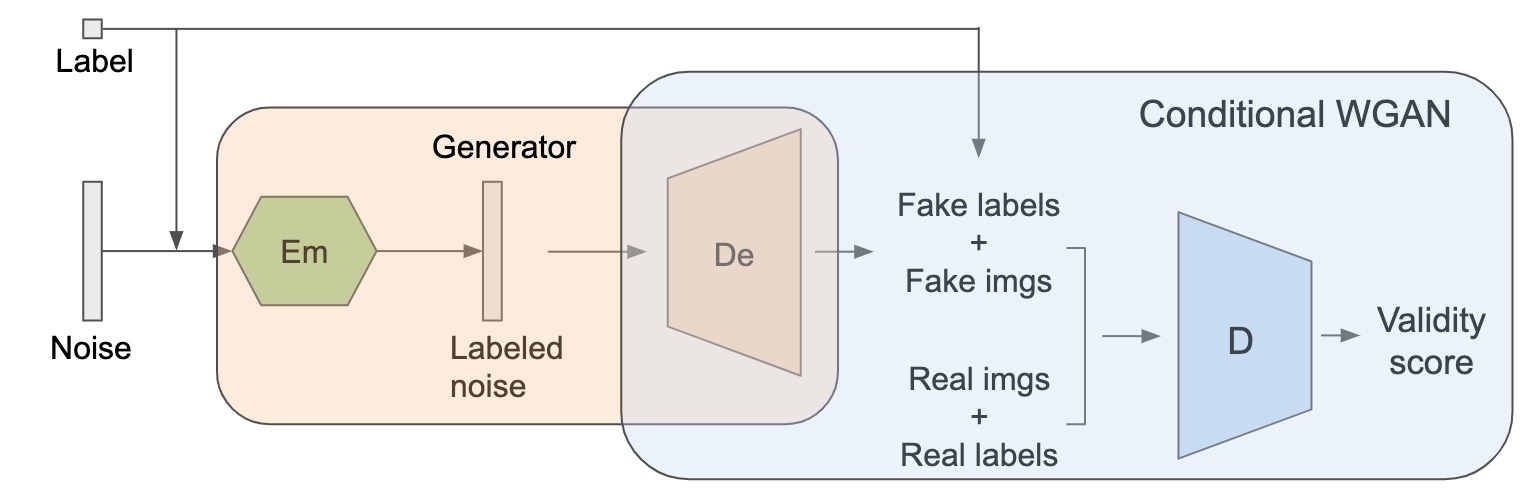}
\caption{\textmd{GAN architecture and our proposed generator.}}
\label{figure-5f2dc111179049b0a2d7f8d132f6c654}
\end{figure}

Our proposed generator is an aggregate model of the pretrained embedding model and decoder model. We feed a random latent vector and a random label into the generator and get a generated image in specific class. The embedding model inside the generator can be updated with GAN training.

\begin{figure}[htbp]
\centering
\includegraphics[width=2.5in, ,height= 1.2in]{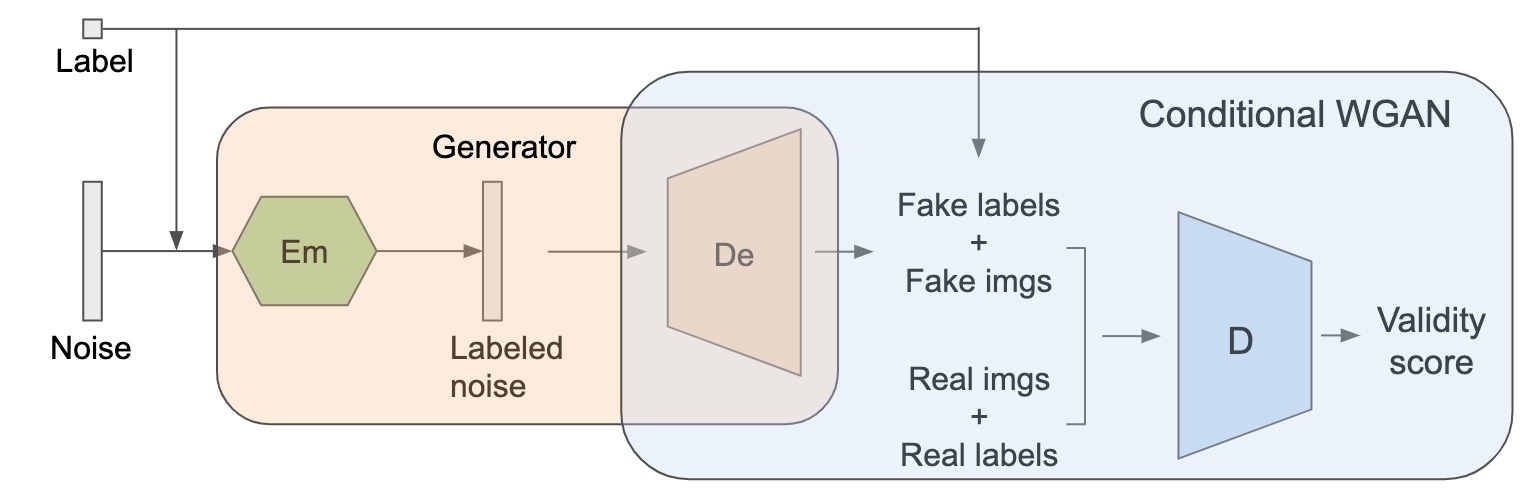}
\caption{\textmd{The discriminator architecture is similar to cWGAN-GP.}}
\label{figure-07f091d8d51d4c949c26d5ffe489065f}
\end{figure}

Our proposed discriminator is an extended model of the pretrained encoder. To note, the discriminator does not use the whole encoder model. Excluding the output layer in decoder, we adopt the second-last output (feature map) and combine the feature map with the embedded labels as a new dense vector. The output of the discriminator is an unconstrained real number, which indicates the total validity of real/fake and class-matching.

\section{Experiments and Results}
The optimizer for our models in this work is Adam algorithm with learning rate 0.0002 and momentum (0.5, 0.9). The size of mini-batches is 128. All the image inputs will be resized as $64\times64\times channels $. The dimension of default latent vector is 128. We only use batch normalization in the generator/decoder. Except the generator's output activation function is \textit{tanh} while the discriminator's is \textit{linear}, other activation functions are \textit{LeakyReLU} with threshold 0.2. Quality of generated images is measured by Fr{\'e}chet Inception Distance. The framework of all experiments is Keras with TensorFlow backend. We use an NVIDIA Tesla P4 GPU with 8GB memory. Most of our results are trained within 3600s. For \textit{Cells} dataset, we train 100 epochs and each epoch takes 18s on our device. For \textit{MNIST Fashion} dataset, we train 15 epochs and each epoch takes 154s on our device. For \textit{CIFAR-10} dataset, we train 30 epochs and each epoch takes 129s on our device.

\textbf{Note.} In each figure of representative images at this section, the first row ($row=0 $) shows real images by class. For each column, we feed the generator with class label $c_{column} $. Start from the second row, we feed the generator with a fixed noise vector $z_{row-1} $. The generated images in this figure are derived by
\let\saveeqnno\theequation
\let\savefrac\frac
\def\dispfrac{\displaystyle\savefrac}
\begin{eqnarray}
\let\frac\dispfrac
\gdef\theequation{16}
\let\theHequation\theequation
\label{dfg-ff888d67a0a9}
\begin{array}{@{}l}Im\left(row>0,column\geq0\right)=G\left(z_{row-1},c_{column}\right)\end{array}
\end{eqnarray}
\global\let\theequation\saveeqnno
\addtocounter{equation}{-1}\ignorespaces

\subsection{MNIST Fashion \& CIFAR-10}

\begin{figure}[htbp]
\centering
\includegraphics[width=3in, ,height= 2in]{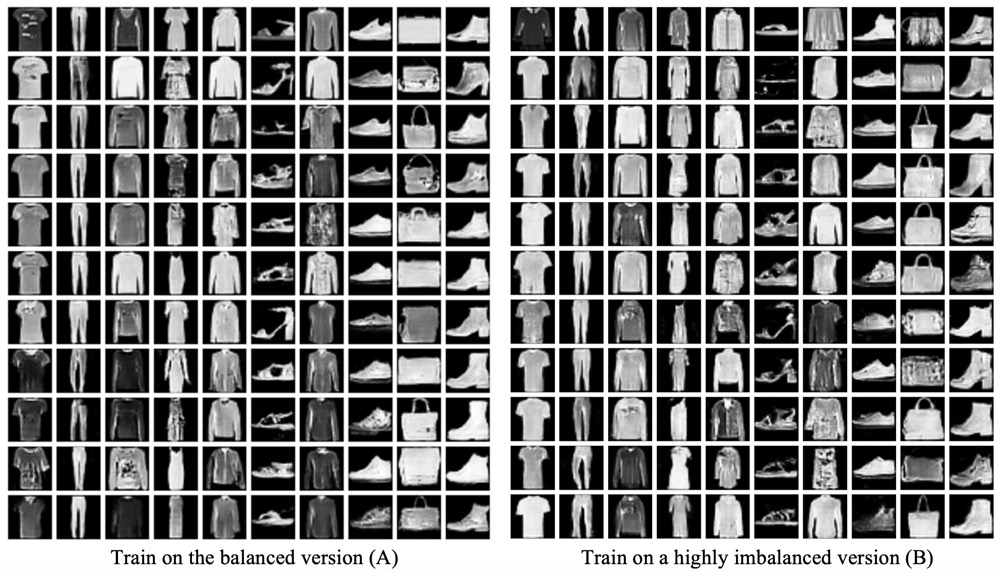}
\caption{\textmd{Representative samples generated in the \textit{MNIST Fashion}. The order of these images follows Equation~(\ref{dfg-ff888d67a0a9}).}}
\label{figure-d8cfe1223c6f40f1af50b5862b7428fc}
\end{figure}

\begin{figure}[htbp]
\centering
\includegraphics[width=3in, ,height= 2in]{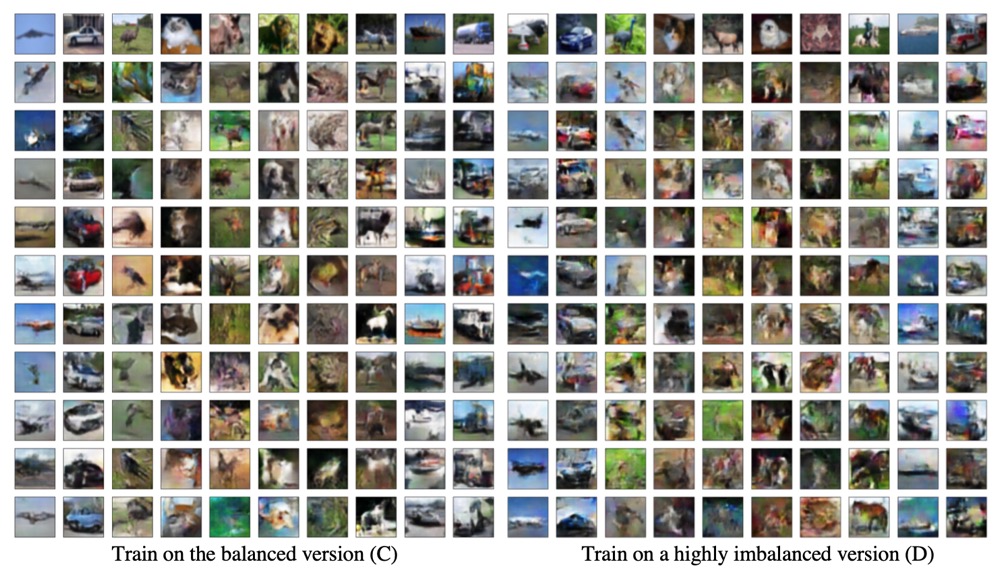}
\caption{\textmd{Representative samples generated in the \textit{CIFAR-10}. The order of these images follows Equation~(\ref{dfg-ff888d67a0a9}).}}
\label{figure-7499e604c14a44c58188ce6b1631ff45}
\end{figure}

\begin{table*}[!htbp]
\caption{{Class weight of \textit{MNIST Fashion} (balanced \& imbalanced)} }
\label{tw-338818b1c283}
\def\arraystretch{1}
\ignorespaces
\centering
\begin{tabulary}{\linewidth}{LLLLLLLLLLL}
\hline
 &
  T-shirt &
  Trouser &
  Pullover &
  Dress &
  Coat &
  Sandal &
  Shirt &
  Sneaker &
  Bag &
  Boot\\
A  &
  4231 &
  4165 &
  4199 &
  4211 &
  4185 &
  4217 &
  4189 &
  4241 &
  4175 &
  4187\\
B &
  4166 &
  73 &
  139 &
  210 &
  287 &
  370 &
  422 &
  387 &
  545 &
  651\\
\hline
\end{tabulary}\par
\end{table*}
We start with our experiments on two well-known balanced datasets, \textit{MNIST Fashion} and \textit{CIFAR-10}. We first sample 70\% of images as the training set for generative models (A for \textit{MNIST Fashion} Table~\ref{tw-338818b1c283}, C for \textit{CIFAR-10} Table~\ref{tw-9c4cb111ee7d}). To exemplify the quality of minority-class generation, an imbalanced version (B for \textit{MNIST Fashion} Table~\ref{tw-338818b1c283}, D for \textit{CIFAR-10} Table~\ref{tw-9c4cb111ee7d}) is created manually for comparison. We observe our model works perfectly not only on the balanced datasets (A, C), but also on the highly imbalanced datasets (B, D). From the representative images Figures~\ref{figure-d8cfe1223c6f40f1af50b5862b7428fc} and~\ref{figure-7499e604c14a44c58188ce6b1631ff45}  generated with imbalanced datasets, we cannot easily figure out which column is minority class. Therefore, our model has a fair training for each class no matter the imbalanced class weight. The learning outcome only depends on the complexity of the image itself. For example, there are 73 \textit{trousers} and 370 \textit{sandals} in dataset B. Although the training set of \textit{sandals} is 5 times as large as trousers, the generated \textit{trousers} images even have a better quality.

\begin{table*}[!htbp]
\caption{{Class weight of \textit{CIFAR-10} (balanced \& imbalanced)} }
\label{tw-9c4cb111ee7d}
\def\arraystretch{1}
\ignorespaces
\centering
\begin{tabulary}{\linewidth}{LLLLLLLLLLL}
\hline
 &
  Airplane &
  Automobile &
  Bird &
  Cat &
  Deer &
  Dog &
  Frog &
  Horse &
  Ship &
  Truck\\
C  &
  3527 &
  3523 &
  3500 &
  3458 &
  3563 &
  3455 &
  3535 &
  3509 &
  3453 &
  3476\\
D &
  3490 &
  71 &
  130 &
  221 &
  269 &
  349 &
  435 &
  485 &
  572 &
  628\\
\hline
\end{tabulary}\par
\end{table*}
The discriminator in our BAGAN-GP has a similar architecture with WGAN-GP. Hence, we can set the train ratio of the discriminator vs the generator to 5 and boost the training with high stability. In the original BAGAN, we cannot set a train ratio larger than 1. Otherwise, the training of BAGAN will be oscillated. In other words, the stability of BAGAN requires a competitive relation between the generator and the discriminator while our BAGAN-GP only pursues a powerful discriminator to lead the generator. Furthermore, our BAGAN-GP still performs excellently when we only initialize the generator because a good generator will accelerate the learning process of the discriminator.

\subsection{Medical image dataset: Cells}\textit{Cells} dataset is a highly imbalanced medical-image dataset, which contains one majority class and three minority classes Table~\ref{tw-5998653fd68c} , i.e. ``red blood cell'', ``ring'', ``schizont'' and ``trophozoite'' respectively. Except the first type, the rest of the cells indicate different stages of malaria infection.

\begin{table}[!htbp]
\caption{{Class weight of \textit{Cells} dataset} }
\label{tw-5998653fd68c}
\def\arraystretch{1}
\ignorespaces
\centering
\begin{tabulary}{\linewidth}{LLLLL}
\hline
 &
  normal \mbox{}\protect\newline (type 0) &
  ring \mbox{}\protect\newline (type 1) &
  schizont \mbox{}\protect\newline (type 2) &
  trophozoite \mbox{}\protect\newline (type 3)\\
Train &
  5600 &
  292 &
  106 &
  887\\
Test &
  1400 &
  73 &
  27 &
  222\\
\hline
\end{tabulary}\par
\end{table}

\begin{figure}[htbp]
\centering
\includegraphics[width=3in, ,height= 2.2in]{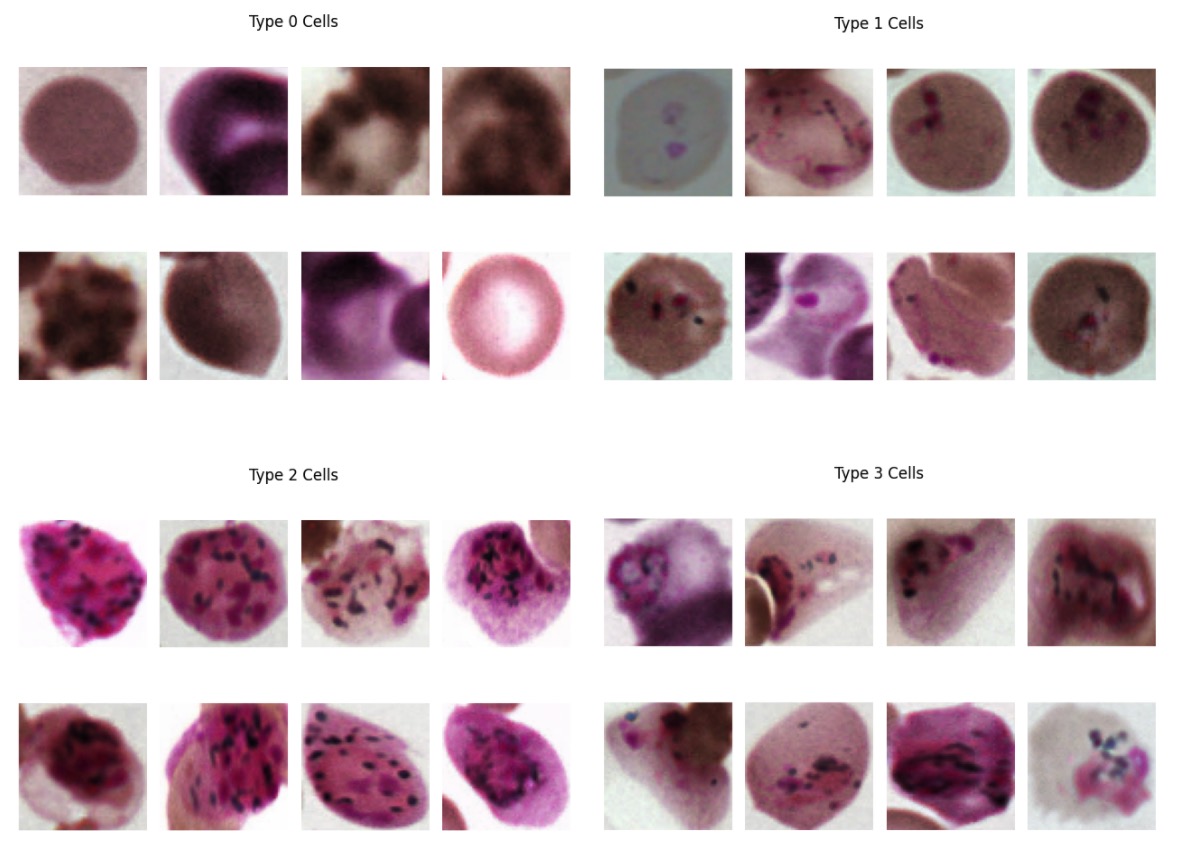}
\caption{\textmd{ Real images per class of \textit{Cells} dataset}}
\label{f-fe972343bfdc}
\end{figure}

Unlike the images of \textit{MNIST Fashion} and \textit{CIFAR-10}, these four classes are different types of red blood cells Figure~\ref{f-fe972343bfdc}. It means they look similar but some different in specific features. Visually, it is hard to distinguish some type 2 cells with type 3 cells.

\begin{figure}[htbp]
\centering
\includegraphics[width=3in, ,height= 2in]{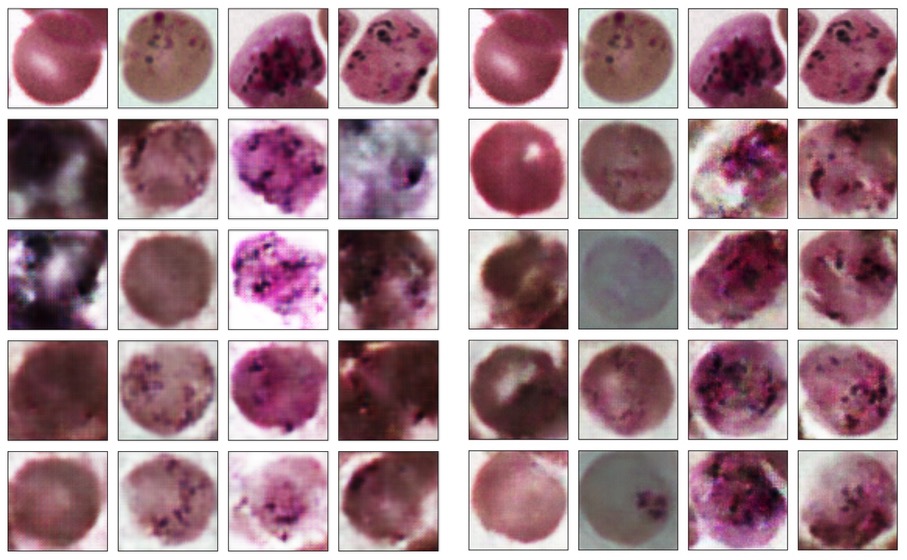}
\caption{\textmd{Generated images by BAGAN (left) and BAGAN-GP (right). The order of these images follows Equation~(\ref{dfg-ff888d67a0a9}).}}
\label{f-3413fa19fe73}
\end{figure}

In Figure~\ref{f-3413fa19fe73} , we observe BAGAN is trying to improve the minority-class generation by sacrificing the quality of majority class. It is exactly the objective of BAGAN, but we are not satisfied on this result. With our BAGAN-GP, all types of cells are generated in high quality. In the section 5, we will quantitatively analyze the performance of our model.

\begin{figure}[htbp]
\centering
\includegraphics[width=3in, ,height= 2in]{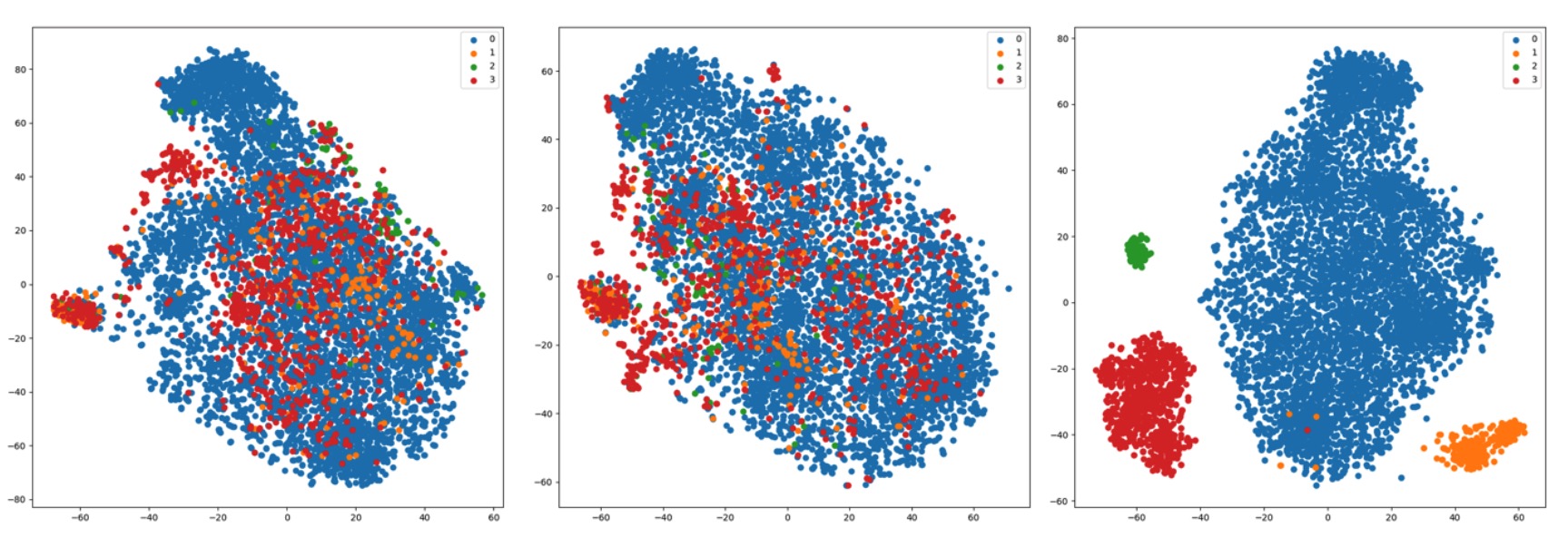}
\caption{\textmd{Two-dimensional t-SNE plot of the encoded latent vectors. Left: Encoder of BAGAN. Middle: Encoder of the improved BAGAN-GP (ours). Right: Encoder + Embedding (ours).}}
\label{figure-1e5fadd157c645edb48385cf30649442}
\end{figure}

In practice, BAGAN is unstable to train on some imbalanced datasets, especially the medical images datasets, e.g. \textit{Cells} dataset in our experiment. The encoder of the original BAGAN cannot translate the input images into dispersed groups of latent vectors Figure~\ref{figure-1e5fadd157c645edb48385cf30649442}. Then, the labeled latent vectors are generated by the distribution of these undivided latent vectors. Thus, the later GAN model will fail to generate images in different classes due to the misleading labeled latent vectors. With our improved autoencoder, we observe that BAGAN becomes stable in training and it is not sensitive to the GAN architecture and hyperparameters.

\begin{figure}[htbp]
\centering
\includegraphics[width=2.5in, ,height= 2in]{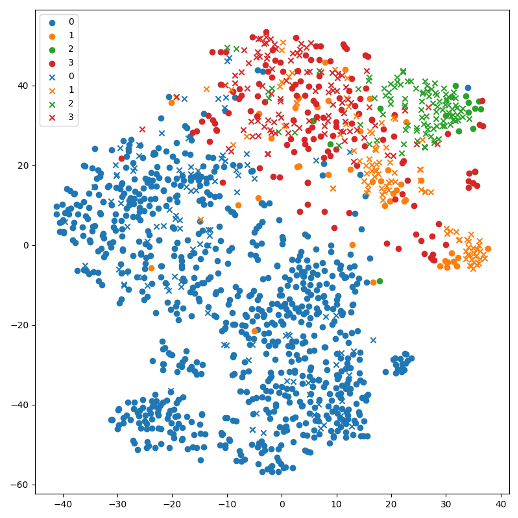}
\caption{\textmd{Comparing the real samples (o) and generated samples (x) by the feature layer output via ResNet-50. }}
\label{figure-b4ca246057024689bd6f4d6d82aa6da6}
\end{figure}

At the feature-level cognition of ResNet-50 Figure~\ref{figure-b4ca246057024689bd6f4d6d82aa6da6}, the generated samples can be regarded as effective augmented images. Furthermore, we observe the generated images manifold are equally distributed around the real images manifold. It means, for each class, our generator is not creating one or few modes of images. In other words, the generator comprehensively learns the real data distribution and does not suffer the problem of mode collapse.

\section{Evaluation}

\begin{table*}[htb]
\caption{{FID: Compare with real samples (validation set)} }
\label{tw-6f6782f573bf}
\def\arraystretch{1}
\ignorespaces
\centering
\begin{tabulary}{\linewidth}{LLLLL}
\hline
 &
  Type0  \mbox{}\protect\newline (1400 samples) &
  Type1  \mbox{}\protect\newline (73 samples)  &
  Type2  \mbox{}\protect\newline (27 samples)  &
  Type3  \mbox{}\protect\newline (222 samples) \\
Rec. samples (Autoencoder) &
  132.715  &
  247.174 &
  322.567  &
  240.519 \\
BAGAN &
  197.961 &
  213.705 &
  278.755 &
  184.903\\
cDRAGAN &
  90.981 &
  184.440 &
  233.512 &
  155.564\\
BAGAN-GP (v1) &
  \textbf{77.831} &
  211.698 &
  227.366 &
  168.240\\
BAGAN-GP (v2) &
  97.445 &
  152.986 &
  213.864 &
  141.798\\
BAGAN-GP (v3, 100) &
  100.151 &
  \textbf{143.703} &
  \textbf{195.926} &
  \textbf{112.875}\\
BAGAN-GP (v3, 200) &
  \textbf{88.562} &
  \textbf{147.994} &
  \textbf{194.544} &
  \textbf{115.881}\\
Real samples (Train)  &
  20.498 &
  93.721 &
  127.392 &
  58.048\\
\hline
\end{tabulary}\par
\end{table*}

There are two common metrics to evaluate the quality of the generated images: Inception Score (IS)\unskip~\cite{850219:19990871} and Fr{\'e}chet Inception Distance (FID) \unskip~\cite{850219:19990877}. Both of these two measurements are based on the Inception V3 network, which is pretrained on ImageNet dataset. IS is derived from the classification logits while FID is derived from the feature layer. IS only measures the distance between the generated sample distribution and the ImageNet distribution, whereas FID calculates the feature-level distance between the generated sample distribution and the real sample distribution. In this work, our objective datasets, medical image datasets, are quite different from ImageNet dataset. Therefore, we adopt FID as the evaluation metric. Fr{\'e}chet Distance is defined as:

\begin{eqnarray*}FID=\ensuremath{\Vert}\mu_r-\mu_g\ensuremath{\Vert}^{2}+Tr\left(\Sigma_r+\Sigma_g-2\left(\Sigma_r\Sigma_g\right)^{1/2}\right)
\end{eqnarray*}
where $\mu_r $ is the mean of the real features, $\mu_g $ is the mean of the generated features, $\Sigma_r $ is the covariance matrix of the real features, $\Sigma_g $ is the covariance matrix of the generated features.

\begin{itemize}
  \item \relax FID on \textit{Cells}. Table~\ref{tw-6f6782f573bf}
\end{itemize}
  All FID scores are calculated by the real samples from validation set and the target samples. For comparison, we introduce two baseline FID scores: the reconstructed samples by autoencoder and the real samples from training set. The FID of reconstructed samples is regarded as a lower baseline and the FID of real samples is regarded as an upper baseline. The quality of target samples is higher when its FID is lower.

In the \textit{Cells} dataset, BAGAN can only generate poor samples. Its performance is only better than autoencoder. As we construct our BAGAN-GP model, we first build a cDRAGAN model Equations~(\ref{disp-formula-group-ec7ed792a3c14dc8b343d2549f6bfa7f}) and~(\ref{disp-formula-group-c84b057286684ce5a018f2db7df4e779})  and combine cDRAGAN with BAGAN framework to get our final model. We need to demonstrate the combined model is better than the previous independent models. cDRAGAN can generate majority-class images with high quality and ignore the minority, which is the drawbacks of non-BAGAN. When we apply autoencoder initialization to cDRAGAN  and keep the same loss function, the BAGAN-GP (v1) can further improve the quality of the majority but there is no improvement on the minority.

\textbf{Note on BAGAN-GP. }(v1): using real labels for generated images Equations~(\ref{disp-formula-group-ec7ed792a3c14dc8b343d2549f6bfa7f}) and~(\ref{disp-formula-group-c84b057286684ce5a018f2db7df4e779}) . (v2): feeding balancing labels in generator at training Equations~(\ref{disp-formula-group-d1a9db7736c7485890c7d108e4251a74}) and~(\ref{disp-formula-group-b72c8c17c6e648ca98ee6b1c48a7d2ce}) . (v3): replacing BAGAN original encoder by our encoder. (100/200): the training epochs. 100 epochs for 1800s, and 200 epochs for 3600s.

Comparing BAGAN-GP (v1) with BAGAN-GP (v2), there is a negative effect on the majority-class generation when we apply balanced training to generator,  which is analogous to BAGAN. However, the improvement on minority-class generation is significant while the negative effect on majority-class generation is small. If our purpose is generating minority-class images, it is recommended to use balanced training (v2). Otherwise, we can omit the balanced training step to generate highest quality images of the majority class. Many traditional GANs will fail to converge with a long training time. Thanks to the gradient penalty term, our BAGAN-GP is stable during a long training period. We observe the longer training on BAGAN-GP, the better overall performance it will achieve.

Although BAGAN-GP is stable with less hyperparameter tuning, here we give some suggestions to build a better BAGAN-GP for future work. In our experiments, we observe it is not recommended to set a high latent dimension and a complex embedding model. Besides, we suggest the discriminator does not need to inherit the weights from the pretrained encoder. The potential reason is the pretrained encoder is not powerful without the embedding part.

\section{Conclusion and Future work}
In this work, we proposed a new architecture of BAGAN with gradient penalty in loss function. With gradient penalty term, we have a more stable BAGAN in training. For the autoencoder initialization, we proposed a supervised autoencoder with an intermediate embedding model to learn the label information directly, which helps to encode the similar but different-class images dispersedly.

We compared the improved BAGAN-GP against the original BAGAN. From the dispersion of labeled latent vectors to the quality of generated images, our model has stronger performance than the original BAGAN. Besides, our model can handle minority class generation in a wide range of datasets, including medical image datasets.

We observe our model can generate images in different classes unambiguously. If we can transfer the class knowledge from generative models to classification models, we believe it will significantly improve the performance of classifiers on imbalanced datasets.

We only use the plain dataset to train the GAN model in this work. In practice, we can apply data augmentation in the step of GAN training, there will be a further improvement on the final results.

There are many research topics dealing with the scarcity of data, such as data augmentation, few-shot and zero-shot learning. We hope our work can broaden the ideas in these topics.

\bibliographystyle{IEEEtran}

\bibliography{article.bib}
\end{document}